\documentclass{article}

\usepackage[final]{neurips_2022}

\usepackage{multirow}
\usepackage{graphicx}
\usepackage{amsmath}
\usepackage{multicol}

\usepackage[utf8]{inputenc} % allow utf-8 input
\usepackage[T1]{fontenc}    % use 8-bit T1 fonts
\usepackage{hyperref}       % hyperlinks
\usepackage{url}            % simple URL typesetting
\usepackage{booktabs}       % professional-quality tables
\usepackage{amsfonts}       % blackboard math symbols
\usepackage{nicefrac}       % compact symbols for 1/2, etc.
\usepackage{microtype}      % microtypography
\usepackage{xcolor}         % colors
\usepackage{caption} 
\title{\texttt{DyREx}: Dynamic Query Representation for Extractive Question Answering}

\author{%
  Urchade Zaratiana$^{1,2}$\thanks{Correspondence to: \texttt{zaratiana@lipn.fr}} , Niama El Khbir$^{2}$, Dennis Núñez$^{2}$, Pierre Holat$^{1,2}$, \\ \textbf{Nadi Tomeh$^{2}$, Thierry Charnois$^{2}$} \\
  $^{1}$FI Group, $^{2}$LIPN, Université Sorbonne Paris Nord - CNRS UMR 7030
}

\begin{document}

\maketitle

\renewcommand{\arraystretch}{1.3}

\begin{abstract}
Extractive question answering (ExQA) is an essential task for Natural Language Processing. The dominant approach to ExQA is one that represents the input sequence tokens (question and passage) with a pre-trained transformer, then uses two learned query vectors to compute distributions over the start and end answer span positions. These query vectors lack the context of the inputs, which can be a bottleneck for the model performance. To address this problem, we propose \textit{DyREx}, a generalization of the \textit{vanilla} approach where we dynamically compute query vectors given the input, using an attention mechanism through transformer layers. Empirical observations demonstrate that our approach consistently improves the performance over the standard one. The code and accompanying files for running the experiments are available at \url{https://github.com/urchade/DyReX}.
\end{abstract}

\section{Introduction}

Extractive question answering is a challenging task where the goal is to extract the answer span given a question and a passage as inputs \citep{SQuAD, NaturalQs}. The prevailing approach achieves Extractive question answerin (ExQA) by firstly producing a contextualized representation of the input, which is a concatenation of the question and the passage, using a pre-trained transformer model. Two learned query vectors are then used to compute a probability distribution over this input sequence representation to produce the start and end positions of the answer span.
This approach has demonstrated very strong and hard-to-beat results, which makes it the de facto approach to extractive QA
\citep{BERT, RoBERTa, span_BERT}. 

However, despite their high performance, we argue that these methods remain suboptimal since the query vectors used to compute the start and end distributions are static, i.e., they are independent of the input sequence, which can be a bottleneck for improving the performance of the model. Hence, we propose to extend this by allowing the queries to dynamically aggregate information from the input sequence to better answer the question. Our method, \textit{DyREx}, iteratively refines the initial query representations, allowing them to aggregate information from the source sequence through attention mechanism \citep{attention_bahdanau, attention_is_all_u_need}. More specifically, we make use of an L-layers transformer decoder architecture, which allows (1) interaction between the queries through self-attention to model the interdependence between the start and end of the answer span, and allows (2) interaction between queries and the input sequence through cross-attention, which specializes the queries to a specific input question and passage, giving more flexibility than a static representation. 

We conduct extensive experiments on several extractive Question Answering benchmarks, including SQuad \citep{SQuAD} and MRQA datasets \citep{MRQA_data}. Experimental results demonstrate that our approach consistently improves the performance over the standard approach.
%%%%%%%%%%%%%%%%%%%%%%%%%%%%%%%%%%%%%%%%%%%%%%%%%%%%%%%%%%%%

\section{Model} 
\subsection{Background: \textit{Vanilla} QA model \label{sec:van}}
We describe here the mainstream approach to extractive Question Answering tasks. In all the following, we call it the ExQA \textit{vanilla} approach. It is typically performed by feeding the input text sequence $\{x_i\}_{i=1}^N$ (the concatenation of the question $Q$ and the passage $D$ containing the answer) into a pre-trained language model such as BERT \citep{BERT}, producing contextualized token representations $\{\textbf{h}_i\}_{i=1}^N \in \mathbb{R}^d$, $d$ being the embedding dimension of the model. Then, to compute the probability of the \textit{start} and \textit{end} positions of the answer span, the following estimators are used:
\begin{align}
    p(\textit{start} = i|Q, D) = \frac{\exp(\textbf{q}_s^T \textbf{h}_i)}{\sum_{i'=1}^{N} \exp(\textbf{q}_s^T \textbf{h}_{i'})} &&
    p(\textit{end} = j|Q, D) = \frac{\exp(\textbf{q}_e^T \textbf{h}_j)}{\sum_{j'=1}^{N} \exp(\textbf{q}_e^T \textbf{h}_{j'})}
\label{eq:1}
\end{align}

Where $\textbf{q}_s$ and $\textbf{q}_e \in \mathbb{R}^d$ are respectively the \textit{start} and \textit{end} queries, randomly initialized and updated during model learning. The training objective is to minimize the sum of the negative log-likelihood of the correct start and end positions $(\hat{i}, \hat{j})$: 
\begin{align}
    \mathcal{L} = - \log p(\textit{start} = \hat{i}|Q, D) - \log p(\textit{end} = \hat{j}|Q, D)
\end{align}

This approach was first proposed by \citet{BERT}, and is now used by most of the work on transformer-based extractive question answering \citep{RoBERTa, span_BERT,shi2022revisiting}. 
% citer les articles

\subsection{Our model: \textit{DyREx}}

The learned query vectors $\textbf{q}_s$ and $\textbf{q}_e$ in the vanilla approach are shared among all sentences and are context insensitive. We presume that using such static queries is a constraining factor for performance improvement, so we propose to extend this approach by allowing the queries to dynamically aggregate information from the input sequence to allow the model to better adapt to the context.

In our model, the initial start and end query representations $\textbf{q}_s^0$ and $\textbf{q}_e^0$ are concatenated and fed to an L-layers transformer decoder \citep{attention_is_all_u_need} to obtain dynamic representations $\textbf{q}_s^L$ and $\textbf{q}_e^L$: 

\begin{equation}
    \textbf{Q}^L = \texttt{Trans\_Dec}_{L}(\textbf{Q}^0, \textbf{H}) 
\end{equation}

with $\textbf{Q}^i = [\textbf{q}_e^i, \textbf{q}_s^i]$ the concactenated queries at layer $i$ and $\textbf{H} = [\textbf{h}_0, \textbf{h}_1, ..., \textbf{h}_N]$ the concatenated token representations, and $\texttt{Trans\_Dec}_{L}$ being an L-layers transformer decoder. 

More specifically, the i-th transformer layer consists of a bi-directional self-attention module $\texttt{self-att}_i$ applied between the queries to model the interdependence between the start and the end positions of the answer, a cross-attention $\texttt{cross-att}_i$ which updates the query representations by aggregating information from the input sequence embeddings, and finally a two-layer point-wise feedforward network $\texttt{FFN}_i$ with GeLU activation \citep{GeLU_act}:
\begin{equation}
    \begin{split}
        \Tilde{\textbf{Q}}^i = \texttt{self-att}_i(\texttt{Q}=\textbf{Q}^i, \texttt{K}=\textbf{Q}^i, \texttt{V}=\textbf{Q}^i) \\
        \widehat{\textbf{Q}}^i = \texttt{cross-att}_i(\texttt{Q}=\Tilde{\textbf{Q}}^i, \texttt{K}=\textbf{H}, \texttt{V}=\textbf{H}) \\
        \textbf{Q}^{i+1} = \texttt{FNN}_i(\widehat{\textbf{Q}}^i)
    \end{split}
\end{equation}

Furthermore, an \texttt{Add-Norm} (skip connection \citep{Resnet} + layer normalization \citep{layernorm}) is inserted after each of the components as in \citet{attention_is_all_u_need}, but we do not show it here for better readability. Moreover, both \texttt{self-att} and \texttt{cross-att} are multi-head scaled dot-product attention from \citet{attention_is_all_u_need}, and the embedding dimension and the number of attention heads of the decoder layers are the same as for the token representation layer.

Finally, to compute the start and the end answer position probabilities, we use the same estimator as the \textit{vanilla} model in equation \ref{eq:1}, substituting $\textbf{q}_s$ and $\textbf{q}_e$ by $\textbf{q}_s^{L}$ and $\textbf{q}_e^{L}$ respectively. Note that the \textit{vanilla} model is a particular case of our model with a number of decoder layers $L=0$.

\section{Experimental setup}
\label{sec:experimental_setup}

\paragraph{Datasets} We evaluate our model on English Machine Reading Comprehension datasets including SQuAD \citep{SQuAD}, HotpotQA \citep{HotpotQA}, TriviaQA \citep{TriviaQA}, NewsQA \citep{NewsQA}, and Natural Questions \citep{NaturalQs}. We preprocess the datasets using standard approaches, then we evaluate our model using the F1 and Exact Match (EM) metrics implemented on the MRQA Github repository.

\paragraph{Hyperparameters} We adopt a similar experimental setup of \citet{few_shot_QA} and \citet{span_BERT}. For all experiments, we use SpanBERT for token representations. We fine-tune our model using hyperparameters from the default configuration of the HuggingFace Transformers library \citep{transformers}. We use Adam optimizer with a learning rate of $3 . 10^{-5}$, where a warm-up stage is set for the first $ 10\%$ of the steps, then decay the learning rate linearly for the rest of the training steps. We employed a batch size of 12, and we trained for a maximum of 5 epochs for full-sized datasets, and for few-shot models, we train the models up to either 2500 steps or 10 epochs. For \textit{DyREx}, unless specified, we employed a 3-layers transformer decoder for the query representations, where each attention layer has 8 heads and the same embedding dimension as the contextualized token embeddings. To run the \textit{vanilla} ExQA model, we borrowed the code from \citep{few_shot_QA} GitHub repository. We trained all the models in a server with V100 GPUs.

% Décrire les datasets

\section{Results}
\label{sec:results}

Table \ref{tab:main_results} presents the obtained results for the different datasets, employing SpanBERT for token representations.
Our approach is very effective in few-shot settings compared to the \textit{vanilla} approach. For instance, our model exceeds the \textit{vanilla} approach by 7.47\% F1 score on SQuAD (512 samples), 5.41\% on HotpotQA (1024 samples), 12.76\% on TriviaQA (256 samples), and 7.97\% on NewsQA (256 samples). These empirical results show the effectiveness of our approach and demonstrate that contextual queries are important when only a few data are available.

\begin{table*}[]
\centering
\begin{tabular}{l|l|ccccc}
\hline
\multirow{2}{*}{Train size} & \multirow{2}{*}{Models} & \multicolumn{5}{c}{Datasets}                 \\ \cline{3-7} 
                          &                         & SQuAD & HotpotQA & TriviaQA & NewsQA & NaturalQs    \\ \hline\hline
\multirow{2}{*}{256} & \textit{Vanilla}                 & 65.74 & 53.23   & 28.49    & 35.80  & 41.87 \\
                          & \textit{DyREx}                    & \textbf{70.75} & \textbf{57.08}    & \textbf{41.66}    & \textbf{43.77}  & \textbf{45.57} \\ \hline
\multirow{2}{*}{512} & \textit{Vanilla}                 & 69.72 & 58.70    & 45.39     & 43.24  & 48.36 \\
                          & \textit{DyREx}                    & \textbf{77.19} & \textbf{61.15}    & \textbf{52.57}    & \textbf{49.20}  & \textbf{55.37} \\ \hline
\multirow{2}{*}{1024} & \textit{Vanilla}                 & 74.01 & 62.54    & 51.87    & 50.61  & 53.42 \\
                          & \textit{DyREx}                    & \textbf{79.42} & \textbf{67.95}    & \textbf{57.59}    & \textbf{54.26}  & \textbf{61.67} \\ \hline
\multirow{2}{*}{Full} & \textit{Vanilla}                 & 90.64 & 79.95    & 76.31    & 68.02  & 77.79 \\
                          & \textit{DyREx}                    & \textbf{91.01} & \textbf{80.55}    & \textbf{77.37}    & \textbf{68.53}  & \textbf{78.58} \\ \hline
\end{tabular}
\caption{\textbf{Main results}. We reported experimental results for different sizes of training datasets using SpanBERT \citep{span_BERT} for token representation.}
\label{tab:main_results}
\end{table*}

\section{Component analysis for \textit{DyREx}}
\label{sec:ablation_study}

We perform a more in-depth analysis to inspect the contribution of the separate components of our \textit{DyREx} architecture. We study the influence of the number of layers of the decoder. We also examine various masking strategies of self-attention of \textit{DyREx} decoder: \textit{Bidirectional} which allows full attention between queries, \textit{Causal} with causal masking (the start query cannot attend the end query), and \textit{Independent} which fully masks the attention between the queries. For all the studies, we employed the SQuAD dataset, a SpanBERT representation, and averaged results across three different seeds.

% @Urchade, in "Ablation Study", this part:
% "So, the influence of \textit{DyREx} decoder layer numbers were studied"
% the part "decoder layer numbers"
% should it be: "the number of decoding layers"?
\begin{figure*}
    \centering
    \includegraphics[width=1.\columnwidth]{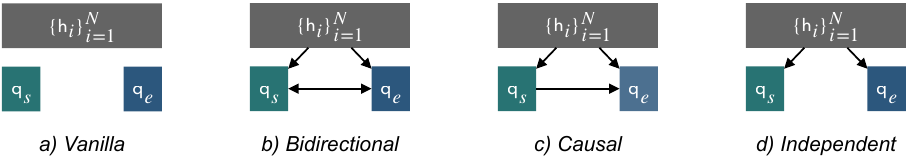}
    \caption{\textbf{Queries and tokens interactions}. This figure shows the different interactions (or attention) between the queries ($q_s$ and $q_e$) and the input tokens ($\{\textbf{h}_i\}_{i=1}^N$). a) \underline{\textit{Vanilla}}: No interaction, either token-query or query-query (i.e., static queries). b) \underline{\textit{Bidirectional}}: dynamic queries with bidirectional query interaction. c) \underline{\textit{Causal}}: dynamic queries with causal relation ($q_s$ influence $q_e$). d) \underline{\textit{Independent}}: queries are independent of each other but remain dynamic.}
    \label{fig:interaction}
\end{figure*}

\begin{table}[]
\parbox{.45\linewidth}{
\centering
\begin{tabular}{c|cc}
\# Layers & F1    & EM    \\ \hline
0          & $90.64 \pm 0.10$ & $83.08 \pm 0.09$ \\
1          & $91.08 \pm 0.05$ & $83.56 \pm 0.31$ \\
2          & $90.92 \pm 0.12$ & $83.50 \pm 0.17$ \\
3          & $91.01 \pm 0.03$ & $83.35 \pm 0.07$ \\
4          & $\textbf{91.17} \pm \textbf{0.05}$ & $83.57\pm0.14$ \\
5          & $91.08 \pm 0.07$ & $\textbf{83.64} \pm \textbf{0.17}$
\end{tabular}
\caption{Results for different number of transformer decoder layers.}
\label{tab:layer_nb_results}}
\hfill
\parbox{.5\linewidth}{
\centering
\begin{tabular}{l|cc}
 & F1    & EM    \\ \hline
\underline{Static} \\
a) \textit{Vanilla} & $90.64 \pm 0.10$ & $83.08 \pm 0.09$ \\
\underline{Dynamic} \\
b) \textit{Bidirectional}       & $\textbf{91.01} \pm \textbf{0.03}$ & $83.35 \pm 0.07$ \\
c) \textit{Causal}      & $91.00\pm0.05$ & $\textbf{83.47}\pm \textbf{0.17}$ \\
d) \textit{Independent}         & $90.87\pm0.09$ & $83.14\pm0.44$
\end{tabular}
\caption{Results for different interactions/attentions between the queries.}
\label{tab:representation_results}}
\end{table}

\paragraph{Influence of \# of layers} Table \ref{tab:layer_nb_results} shows that performance is generally better with more layers. The best performance is attained when using 4 or 5 layers, but a lower number of layers can also obtain competitive results. However, the results are much weaker without a decoder layer, i.e., without a static query representation.

\paragraph{Type of self-attention} Table \ref{tab:representation_results} shows the obtained results using the various attention or masking strategies for the decoder. We see that fully masking (d) attention between queries provides the weakest result, while \textit{Causal} and \textit{Bidirectional} attention perform similarly. This shows that the interaction between queries is important.

\section{Related Work}
\label{sec:related_work}

Early ExQA models based on deep learning, such as BidAF \citep{bidaf}, Match-LSTM \citep{Match-LSTM}, QaNet \citep{QaNet} and others, were highly specialized and heavily engineered. The arrival of BERT and pre-trained language models completely transformed the domain by introducing a simple yet effective approach: the \textit{Vanilla} ExQA (section \ref{sec:van}) \citep{BERT}. Since its introduction, this has been the dominant approach to extractive QA \citep{RoBERTa, span_BERT}. For instance, \citep{Fajcik2021RethinkingTO} proposes to model the joint probability of the spans, instead of modeling the probability of a span’s start and end positions independently, but it does not outperform the \textit{Vanilla} approach. In this work, we extend the vanilla model not by modifying the objective function but by learning a richer representation of the query parameters.

\section{Conclusion}
In this paper, we propose \textit{DyREx}, a method to dynamically compute query representations to calculate the start and end positions of answer spans in extractive question answering. Our approach consistently outperforms the dominant approach on a wide range of QA datasets, and the gain is even more significant in a few-shot scenario. In future work, it would be interesting to adapt \textit{DyREx} for multi-span extraction tasks such as Named Entity Recognition and Keyphrase Extraction.

\section*{Acknowledgments}
 This work is partially supported by a public grant overseen by the French National Research Agency (ANR) as part of the program Investissements d’Avenir (ANR-10-LABX-0083). This work was granted access to the HPC/AI resources of [CINES/IDRIS/TGCC] under the allocation 20XX-AD011013682 made by GENCI.

\bibliography{custom.bib}
\bibliographystyle{plainnat}
\appendix

\end{document}